\documentclass[preprint,12pt]{elsarticle}

\usepackage{amssymb}

\usepackage{subcaption}
\usepackage{multirow}
\usepackage{algorithm,algpseudocode}
\usepackage{xcolor}
\usepackage{soul}
\usepackage{amsmath}

\journal{Expert Systems with Applications}

\begin{document}

\begin{frontmatter}

\title{LM-IGTD: a 2D image generator for low-dimensional and mixed-type tabular data to leverage the potential of convolutional neural networks}

\author[label1]{Vanesa Gómez-Martínez}\ead{vanesa.gomez@urjc.es}
\author[label1]{Francisco J. Lara-Abelenda}\ead{francisco.larag@urjc.es}
\author[label1]{Pablo Peiro-Corbacho}\ead{pablo.peiro@urjc.es}
\author[label1]{David Chushig-Muzo}\ead{david.chushig@urjc.es}
\author[label2]{Concei\c{c}\~{a}o Granja}\ead{conceicao.granja@ehealthresearch.no}
\author[label1]{Cristina Soguero-Ruiz}\ead{cristina.soguero@urjc.es}

\affiliation[label1]{organization={Department of Signal Theory and Communications, Telematics and Computing Systems, Rey Juan Carlos University},
             %addressline={Madrid},
             city={Madrid},
             postcode={28943},
             %state={},
             country={Spain}
             }

\affiliation[label2]{organization={Norwegian Centre for E-health Research, University Hospital of North Norway},
             city={Troms\o},
             postcode={9008},
             country={Norway}
             }

\begin{abstract}
Tabular data have been extensively used in different knowledge domains. Convolutional neural networks (CNNs) have been successfully used in many applications where important information about data is embedded in the order of features (images), outperforming predictive results of traditional models. Recently, several researchers have proposed transforming tabular data into images to leverage the potential of CNNs and obtain high results in predictive tasks such as classification and regression. In this paper, we present a novel and effective approach for transforming tabular data into images, addressing the inherent limitations associated with low-dimensional and mixed-type datasets. Our method, named Low Mixed-Image Generator for Tabular Data (LM-IGTD), integrates a stochastic feature generation process and a modified version of the IGTD. We introduce an automatic and interpretable end-to-end pipeline, enabling the creation of images from tabular data. A mapping between original features and the generated images is established, and post hoc interpretability methods are employed to identify crucial areas of these images, enhancing interpretability for predictive tasks. An extensive evaluation of the tabular-to-image generation approach proposed on 12 low-dimensional and mixed-type datasets, including binary and multi-class classification scenarios. In particular, our method outperformed all traditional ML models trained on tabular data in five out of twelve datasets when using images generated with LM-IGTD and CNN. In the remaining datasets, LM-IGTD images and CNN consistently surpassed three out of four traditional ML models, achieving similar results to the fourth model.

\end{abstract}


\end{frontmatter}


\section{Introduction}
\label{}

The steady increase in computational resources and availability of large datasets have accelerated the success of deep learning (DL)~\cite{borisov2022deep}. DL has marked a milestone in modern society, leading to great success in various domains (computer vision, natural language processing, and autonomous systems) and a plethora of studies such as image recognition, speech recognition, machine translation among others~\cite{oord2016wavenet, devlin2018bert, he2016deep, lauriola2022introduction}. Among DL-based models, the convolutional neural networks (CNN) have proven high predictive performance in image applications such as segmentation~\cite{zhang2015deep}, registration~\cite{wu2015scalable}, fusion~\cite{suk2014hierarchical}, and classification~\cite{krizhevsky2017imagenet}. The capability to exploit spatial relationships in data and the great predictive performance of CNN-based models has substantially improved the results of multiple computer vision tasks and contributed to other fields, including healthcare~\cite{anwar2018medical}. Although CNNs perform exceptionally well with images, the application of CNN-based models to tabular data remains challenging because of the distribution of data (samples, features) and lack of spatial information~\cite{arik2021tabnet, medeiros2023comparative}.

In many fields, including industry, finance, medicine, and climate science, tabular datasets are ubiquitous~\cite{shwartz2022tabular}. Traditional ML, particularly tree-based models, have dominated the results and benchmark the state-of-the-art when tabular data are used, showing superior performance to DL-based models~\cite{grinsztajn2022tree, bragilovski2023tltd}. Owing to a growing interest in leveraging the high predictive performance of DL-based models, specially CNNs, there has been an increasing research interest focused on transforming tabular into visual representations, \textit{e.g.,} 2D and 3D images. Several researchers have proposed methods to transform tabular data into images~\cite{wolf2022daft, lee2022tab2vox, sharma2019deepinsight, bazgir2020representation, zhu2021converting}. The idea behind the algorithms is to transform samples of tabular data into images and then utilize them as input for CNN-based models, leveraging the high predictive capability for tabular data. %Moreover, it can be used as a feature extractor by extracting informative information and leading to better interpretability. 
Additionally, it can be applied as a feature extractor, improving interpretability by extracting useful information. Thanks to the ongoing research on CNN, several interpretability techniques can be applied, such saliency map visualization (\textit{e.g.,} Grad-CAM~\cite{zhou2016learning}), which can enhance both the prediction accuracy and the model interpretability.

Although the models for transforming tabular data to images have proven to be effective in different applications, results showed that their performance was excellent in \textit{high-dimensionality} scenarios \textit{e.g.,} with a large number of features. When CNNs are applied to tabular data, several challenges arise, mainly data sparsity (thereby, missing values), low-dimensionality, and the presence of different types of features~\cite{shwartz2022tabular}. The presence of missing data results in images with scarce information, with white tones and many blanks unrelated to any feature~\cite{damri2023towards}, producing worse image quality and hampering their use in CNNs. The number of features (\textit{high-dimensionality}) is a key factor in the performance of most tabular-to-image methods because of is directly related to image size. To properly train CNN-based models, images require images of medium and large sizes~\cite{hashemi2019enlarging}. Recent works related to tabular data benchmarks also ignore datasets composed of mixed-type data, \textit{i.e.,} with numerical and categorical features. To the best of our knowledge, this is the first study that considers the transformation of tabular data with mixed numerical and categorical data into 2D images. Therefore, achieving the implementation of a method capable of transforming these tabular datasets into images, and subsequently using them in CNN-based models, would represent a significant advancement and improvement in predictive performance. 

% Contributions
The main contributions of this paper are summarized as follows. 
\begin{enumerate}
    \item A novel and effective approach designed to create synthetic and noisy features (\textit{noise data generation}) that address the inherent limitations of transforming tabular data to images from low-dimensional and mixed-type datasets.
    \item An automatic and interpretable end-to-end pipeline that combines \textit{noise data generation}, and a modified version of Image Generator for Tabular Data (IGTD)~\cite{zhu2021converting} method to convert tabular data to 2D images called Low Mixed-IGTD (LM-IGTD). A mapping between original features and images created is built, and post hoc interpretability methods are applied to identify salient areas of these images and gain interpretability for classification tasks.
    \item An extensive evaluation of the tabular-to-image generation approach proposed on 12 low-dimensional and mixed-type datasets, including binary and multi-class classification scenarios. Several predictive ML models are used as benchmarks and used to compare the results of using images and CNN-based models. 
\end{enumerate}

The rest of the paper is organized as follows. In Section~\ref{sec:related_work}, we review in-depth the methods that transform tabular data into images in the literature. The methods used and a further description of the methodology proposed is detailed in Section~\ref{sec:methods}. In Section~\ref{sec:results}, we present the experimental setup, and the results for binary classification and multiclass classification. Additionally, we conduct an interpretability process on the images generated and apply post-hoc interpretability methods on trained CNN-based models. The discussion and conclusions are shown in Section~\ref{sec:discussion} and Section~\ref{sec:conclusion}, respectively.

\section{Related work}
\label{sec:related_work}

Recently, several authors have developed methods that convert samples of tabular data into images and subsequently using them as input to CNN-based models~\cite{wolf2022daft, lee2022tab2vox, sharma2019deepinsight, bazgir2020representation, zhu2021converting}. We have categorized the tabular-to-image methods into three: feature permutation-based, dimensionality-reduction (DR)-based and embedded-based methods. The feature permutation-based methods determine the permutation of features to assign feature values to compute pixel values of the created image. The feature permutation-based methods, determine the permutation by minimizing the errors between the feature distance rank and pixel distance rank. DR-based methods use nonlinear methods that maintain the topology of input features to a 2D plane and preserve the local relationships among features following the similarity among them in the input space.

The DR-based methods identified in the literature were DeepInsight, REFINED, DeepFeature, Vec2image, DCNNTr, TINTO, FC-Viz, Fotomics. A further detail of these methods are detailed as follows. DeepInsight used manifold DR methods such as t-SNE and kPCA with different distances (\textit{e.g.,} Euclidean, Mahalanobis, and Chebychev) for transforming high-dimensional tabular data to images~\cite{sharma2019deepinsight}. After DR, they find a minimum area rectangle that contains the projected samples using the convex Hull algorithm, and convert the coordinates to pixels, mapping the feature values to their respective pixel locations. Several additional steps to enhance the transformation process such as rotation of Cartesian coordinates, determination of pixel locations, and mapping of elements to their respective pixel locations. DeepInsight has been utilized across various applications, including single-cell datasets, Alzheimer's data, and artificial datasets~\cite{jia2023scdeepinsight}. DeepInsight allows the overlap of feature positions and calculates the average of them as pixel values.

REFINED~\cite{bazgir2020representation} algorithm is based on four stages: (1) a mapping process where to each feature is assigned a unique pixel using the Euclidean distance; (2) the creation of an initial feature map that preserves the feature distances into a 2-D space using the DR method called multidimensional scaling (MDS); (3) to estimate the location of the feature on a bounded domain the Bayesian MDS is applied, with the constraint that each pixel could contain at most one feature; and finally, the hill-climbing algorithm is used to optimize and compute the pixel positions in the image~\cite{bazgir2020representation}. REFINED modifies the local position after DR to avoid the overlap.

EDLT method computed a pairwise feature-feature correlation matrix $\mathbf{M}_c$ and feature-label correlation vector $\mathbf{l}_c$ using Pearson correlation~\cite{han2019convolutional}. The first aimed to capture the similarity among features, whereas the second to evaluate the relevance of each feature to the class label. Then, EDLT constructs a feature reordering matrix $\mathbf{O}$ using $\mathbf{M}_c$ and $\mathbf{l}_c$. Finally, by using $\mathbf{O}$, EDLT reorders original feature values and converts each sample into a synthetic matrix (2D image representation), where feature values in adjacent rows and adjacent columns share spatial correlation resembling to the spatial adjacent areas in an image.

DeepFeature~\cite{sharma2021deepfeature} used the same algorithm as DeepInsight~\cite{sharma2019deepinsight} for image transformation. According to~\cite{sharma2021deepfeature}, when an image is transformed from Cartesian coordinates to pixel coordinates, it brings distortion due to the limited or fixed size of the pixel frame and many features may overlap on the same location~\cite{sharma2021deepfeature}. The non-quantized compression algorithm called the snowfall compression algorithm is included to find nearby empty pixel locations of features and reduce overlapping. 

Vec2image uses nonlinear DR methods (t-SNE, k-PCA and UMAP) to transform the feature values from the input space into a two-dimensional representation (2D image). Next, a rotation (horizontal or vertical) is performed by a convex hull algorithm to frame the 2D image as input to CNN-based models, and the Cartesian coordinates are converted to pixels, thus determining the location of each feature in the pixel frame. Finally, the expression values of each sample within the vector are mapped to these pixel locations as detailed pixel values. Vec2image uses SMOTE to address the class imbalance and improves classification performance when trained on imbalanced data.

In Wrangling~\cite{sharma2022classification}, authors proposed three data wrangling techniques to convert 1-D vectors to 2-D image representations. Specifically, equidistant bar graph, normalized distance matrix, and the combination of both. The first two methods produce binary and grayscale images, respectively. The third method produces a colored 2-D image with 3 layers: (1) it contains the normalized distance matrix; (2) it has bar graphs; and (3) it has a copy of numerical data stored row-wise. These algorithms were only validated using tabular data associated with breast cancer, specifically the Wisconsin breast cancer Wisconsin diagnostic breast cancer.

% The deep CNN-based transfer learning 
DCNNTr~\cite{sisodia2022feature} method transforms a 1-D vector into a 2-D image data through the following steps: (1) dimensionality reduction and element arrangement using t-SNE; (2) convex hull using Graham’s algorithm; and (3) image alignment.

DWTM uses Pearson correlation and chi-square to compute a weight for each feature~\cite{iqbal2022dynamic}. The features are arranged in descending order according to these computed weights. Next, each feature is assigned a portion of space in the image canvas (canvas size allocation) based on the value of their corresponding weights. DWTM calculates the length, height, and area required for each feature by using the ratio of the weights of each feature to the sum of the total weights of all features and distributes the image canvas space accordingly. Several features have a long floating-point sequence, to avoid the creation of a big image a portion of the sequence after the floating-point as are embedded into the canvas, and the rest is trimmed. DWTM ensures the maximum utilization of the image canvas space and allows to create of images in multiple sizes~\cite{iqbal2022dynamic}.

TINTO is based on four stages: (1) DR with PCA and t-SNE methods; (2) determining the center of gravity of the points and the area is subsequently delimited; (3) scaling and pixel positions, where the matrix is transposed, scaled and the values are rounded to integer values; and (4) the values obtained would be the positions of the characteristic pixels for the creation of the image pattern.

FC-Viz transforms tabular data into 2D images using four stages: (1) finding clusters of highly correlated features and measuring interactions between representative features of each cluster~\cite{damri2023towards}; (2) using an optimization algorithm to find an optimal spatial arrangement of feature clusters; and (3) applying DR techniques are used to determine the value of each pixel~\cite{damri2023towards}. 

Fotomics uses an image transformation method to convert non-image omics data into images using fast Fourier transform (FFT) to map features onto a two-dimensional Cartesian plane (2D image representation). Then, a convex hull algorithm is used to find the smallest convex hull that contains the samples within the Cartesian plain, followed by rotation to frame the plane vertically or horizontally to comply with the CNN input orientation. The proposed method was applied to single-cell RNA-seq data for cell-type classification using CNN models.

HACNet is an end-to-end learning model consisting of two components, an attention-based table-to-image converter and a CNN-based predictor~\cite{matsuda2023hacnet}. The converter creates an image by calculating each pixel value through an attention mechanism and calculates the pixel values by the inner products between attention weights and feature values. The Gumbel-Softmax with decreasing temperature is used to generate attention weights in the training phase and become one-hot vectors in the prediction phase. This one-hot vector implements a feature selection scheme. The converter parameter and the weights of CNN are trained simultaneously in the first phase, and the weights of CNN are additionally trained in the second stage. 

% Binary Image Encoding (BIE)

The BIE~\cite{briner2023tabular} method converts network flow data into 2D images. First, BIE transforms each numerical sample value into a 64-bit floating point representation (a bit representation that takes `0' and `1' values). Then, these representations are stacked on top of each other to create a 2-D image image, with zero and one values represented by black pixels and white pixels, respectively. BIE does not use normalization, which can reduce the range of potential feature values and diminish the impact of outliers.

An overview of methods for transforming tabular data into images is shown in Table~\ref{table:summary_tabular_image_methods}, detailing the application, domain, number of samples, and number of features where these methods have been employed. 

\begin{table}[!ht]
\centering
\caption{Overview of the methods proposed in the literature for transforming tabular data into images. The application, domain, number of samples, and the number of features where these methods were applied are shown.}
\label{table:summary_tabular_image_methods}
\scalebox{0.85}{
\begin{tabular}{|l|l|l|l|}  \hline
Method                                              & Type  & Domain/       & Type       \\ 
                                                    & Image & Application   & method     \\ \hline

\textsl{DeepInsight}~\cite{sharma2019deepinsight}   & 2D image  & RNA-sequences, vowels,    & DR-based  \\ 
                                                    &           & synthetic data    & \\\hline

\textsl{REFINED}~\cite{bazgir2020representation}    & 2D image  & clinical                  & DR-based  \\
                                                    &           &     & \\ \hline

\textsl{EDLT}~\cite{han2019convolutional}           & 2D image  & benchmark datasets        & feature permutation-based  \\ 
                                                    &           &                           &  \\ \hline

\textsl{DeepFeature}~\cite{sharma2021deepfeature}   & 2D image  & RNA-seq data              & DR-based   \\ 
                                                    &           &  & \\\hline                                         

\textsl{IGTD}~\cite{zhu2021converting}              & 2D image  & Clinical data             & feature permutation-based   \\ 
                                                    &           & &\\ \hline

\textsl{Vec2image}~\cite{tang2022vec2image}         & 2D image  & RNA sequencing data       & DR-based  \\ 
                                                    &           & &\\ \hline

\textsl{Wrangling}~\cite{sharma2022classification}  & 2D image  & benchmark datasets & embedded-based   \\ 
                                                    &           & &\\ \hline

\textsl{DCNNTr}~\cite{sisodia2022feature}           & 2D image  & Detection of click fraud & DR-based \\ 
                                                    &           & &\\ \hline

\textsl{DWTM}~\cite{iqbal2022dynamic}               & 2D image  & six benckmark datasets  &  embedding-based \\ 
                                                    &           & & \\ \hline

\textsl{TINTO}~\cite{castillo2023tinto}             & 2D image  & Commercial  & DR-based \\ 
                                                    &           & &\\ \hline

\textsl{FC-Viz}~\cite{damri2023towards}             & 2D image  & bechmark high-dimensional & DR-based  \\ 
                                                    &           & datasets &\\ \hline

\textsl{Fotomics}~\cite{zandavi2023fotomics}        & 2D image  & single-cell RNA-seq data & DR-based  \\ 
                                                    &           &  & \\ \hline

\textsl{HACNet}~\cite{matsuda2023hacnet}            & 2D image  & gene expression, synthetic & end-to-end-based   \\ 
                                                    &           & data &\\ \hline

\textsl{BIE}~\cite{briner2023tabular}               & 2D image  & network flow data & embedding-based   \\ 
                                                    &           &&\\ \hline

\end{tabular}
}
\end{table}

\section{Methods}
\label{sec:methods}

\subsection{Convolutional neural networks based models and post-hoc interpretability}
\label{sec:convolutional_networks}

CNN-based models have demonstrated exceptional performance computer vision applications, achieving outstanding results specially in image classification~\cite{li2021survey}. CNNs effectively extract spatial information through convolution operations, which contributes to great success in applications where information rely on the order of the features (\textit{e.g.,} images, audio)~\cite{li2021survey}. Despite the high predictive capability of CNNs and excellent results in various domains, the use of nonlinear transformations during the training result in models with a lack of interpretability~\cite{yang2022unbox}. Recently, explainable artificial intelligence (XAI) methods has emerged to provide interpretability and show how the model's predictions are reached, thus enhancing the transparency and trustworthiness on trained models~\cite{vilone2021notions, yang2022unbox}. To increase the interpretability of CNNs and image applications, several methods have been proposed such as attribution methods such as heatmaps, saliency maps or class activation methods. Among them, class activation maps (CAM) are XAI methods that helps in the visualization and localization of salient areas in images, leading to support decision-making process, specially in medical image analysis~\cite{yang2022unbox}. CAM is a technique for identifying discriminative regions by linearly weighted combination of activation maps of the last convolutional layer before the global pooling layer. GradCAM~\cite{selvaraju2017grad}, an extension of CAM, is a local post-hoc interpretability methods that uses the global average of the gradient to determine the importance of feature map~\cite{zhou2016learning}. The CNN layers are well-known for capturing both spatial information and high-level semantics. With this foundation in place, the final CNN layer may have the optimal composition for extracting important data~\cite{ali2023explainable}. Grad-CAM assigns significance ratings to each neuron for the given target class using the gradient information backpropagated to the final convolutional layer. Grad-CAM was selected because it can be applied to any range of CNN-based models without needing to go through architectural modifications or retraining, and its effectiveness has proven in multiple studies~\cite{selvaraju2017grad}. 

\subsection{Stochastic noise generation for augmenting dimensionality with mixed-type features}
\label{sec:noise_generation}

% Why noise generation
Noise is usually regarded as a unfavorable factor for research because it can represent uncertainty in data~\cite{chen2022noise}. Nevertheless, in the state-of-the-art, several researchers have explored the use of noise in different applications. By adding noise to create noisy variants of the original images and enhance the generalization of CNNs and improve prediction results~\cite{momeny2021learning}. The noise injection has also shown to be useful as regularization method to improve robustness of latent representations~\cite{vincent2010stacked, chushig2022learning}. Inspired by data augmentation techniques, we aggregate noisy features to input data to extend the dimensionality of tabular data, and then create images with medium size. Our hypothesis is that the inclusion of noisy features can help to create \textit{new synthetic pixels} in the image generated helping to capture intrinsic relationships between original features, and the spatial relationship of an image. To address the limitations of most of tabular-to-image methods, we propose the creation of noisy features for feature data augmentation, thus improving the images generated and the subsequent classification performance. In this study, the generation of noise is characterized by three parameters: \textit{(i)} the noise type; \textit{(ii)} the noise power to apply; and \textit{(iii)} the number of noisy features that be created regarding the original features.

% Types of noise
To create noisy variables, we evaluated with four noise types: Gaussian, swap noise (SWN), zero-masking (ZMN), salt-and-pepper noise (SPN). Gaussian noise is extensively used in many applications, but it can be applied to numerical features. As stated previously, mixed-type datasets comprise both numerical and categorical features and to handle correlations for these features,  SWN, ZMN and SPN were used. SWN overwrites the value of a category with another value randomly sampled from the same feature. For example, let a categorical feature $\mathbf{f}_i$ with three categories: a, b, c. Considering that a sample $\mathbf{x}$ has an in $\mathbf{f}_i$, we replace its value with another value, such as b or c. Power noise controls the number of categorical values corrupted in $\mathbf{f}_i$. ZMN and SPN corrupt a fraction of the feature (controlled by the noise power), setting the value to the minimum in ZMN and setting the value to the minimum/maximum value (according to a fair coin flip) in the SPN. ZMN and SPN were used to measure relationships between binary features, whereas SWN for categorical with more than 2 categories.

% Homogeneous and heterogeneous noise. Feature selection.
In our approach, we distinguish two types: homogeneous noise generation (HoNG) and heterogeneous noise generation (HeNG). For HoNG, we create a similar number of noisy features for each feature in the input dataset, whereas for HeNG, the number of noisy features associated with each feature depends on the feature importance obtained by feature selection (FS) methods. FS methods seek to identify a subset of features that capture the underlying information of the input data, whereas simultaneously discard irrelevant features~\cite{remeseiro2019review}. FS lead to lower computational complexity and improving both generalization capacity and performance in subsequent predictive models~\cite{remeseiro2019review}. FS methods are mainly categorized into three categories, filter, wrapper and embedding methods~\cite{pes2020ensemble}. The filter methods focus on exploring the intrinsic properties of data by using some scoring functions to rank the given features. Filter FS methods assign an importance value to each input feature, which represents how much important is for a predictive task. To enhance the performance of traditional FS methods, ensemble FS has been lately studied~\cite{pes2020ensemble}. The core idea behind ensemble learning is to combine multiple \textit{base learners} to solve a specific problem. In ensemble FS, the goal is to combine the selected features by \textit{base FS learners} to produce a robust and stable selection of features~\cite{seijo2017ensemble}. The hipothesis is that if the FS is repeated using slightly different training data, the frequency of the most important features chosen will be high, whereas irrelevant features are less frequently selected~\cite{pes2020ensemble}. In this study, for HeNG, we used ensemble FS with filter methods (Relief, mRMR) to identify the most relevant features and generate more noisy features to them. To determine the number of features, the type of noise, and noise power, two algorithms based on HoNG and HeNG were developed. 

To validate the similarity between real and noisy features generated, several correlation measures were used. Since mixed-type features are considered the selection of the correlation depends on the type of feature. Spearman correlation (SPC) was selected to quantify linear relationship between numerical features~\cite{yu2022robust}. Point-biserial correlation (PBC) measures the linear correlation between binary and continuous variable~\cite{kornbrot2014point}. Finally, phik correlation allows to measure the relationship between mixed-type features (categorical and numerical variable)~\cite{baak2020new}.

\subsection{IGTD for low-dimensional and mixed-type datasets}
\label{sec:igtd}

IGTD transforms tabular data into grayscale 2D images. It assigns one pixel in a grayscale image for each input feature and the intensity reflects the value of the feature~\cite{zhu2021converting}. IGTD calculates the distances of each feature to every other (pairwise distance) and attempts to assign pixels with similar distances in the image. Towards that end, it computes of the distance matrix of the features (\textit{feature ranking matrix}) and pixels (\textit{pixel ranking matrix}) in ascending order (ranked). The IGTD optimizes the assignment of features by minimizing the differences between the two matrices. The IGTD algorithm seek to ensure that similar features are placed close to each other and dissimilar ones further away from each other in the image.

In the original paper~\cite{zhu2021converting}, pairwise distances were calculated through the Euclidean distance and the Pearson correlation. The efficiency of IGTD is based on a greedy iterative process of swapping the pixel assignments of features to reduce the distance between them. At each iteration, the algorithm determines the feature that was not considered for swapping and looks for a feature to replace it that will minimize the gap between the two ranking matrices (\textit{feature ranking matrix} and \textit{pixel ranking matrix}). One may alternatively construct a dissimilarity measure that can be applied to mixed data. Typically, such a dissimilarity measure can be constructed by defining and combining dissimilarity measures for each type of variable.

\begin{figure*}[h]
\centering
\begin{subfigure}[b]{\linewidth}
\includegraphics[width=\textwidth]{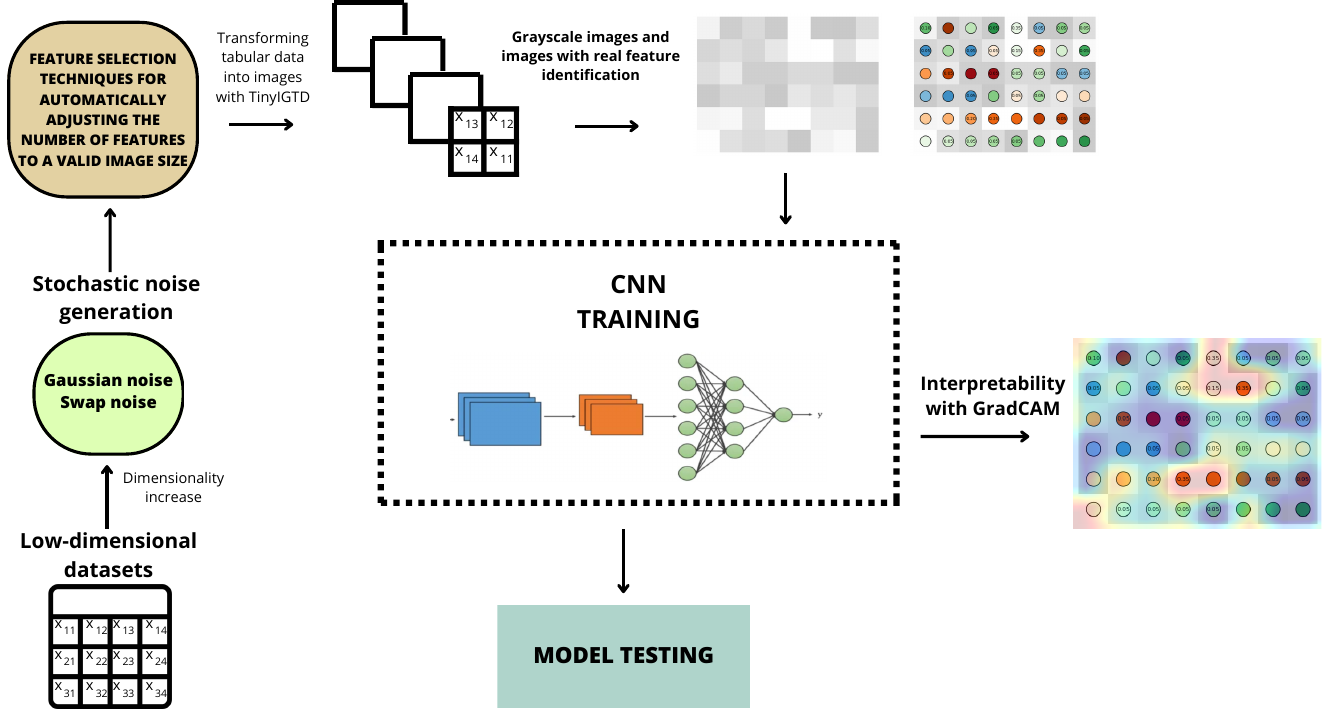}
\end{subfigure}
\caption{Workflow of proposed methodology.}
\label{fig:workflow}
\end{figure*}

\section{Results}
\label{sec:results}

In this section, we present the experimental results by comparing ML-based models and our tabular-to-image methodology in datasets with low-dimensional and mixed-type features. Furthermore, we conduct an ablation study to demonstrate the impact of the homogeneous and heterogeneous approaches for noise generation in our proposed approach. 

\subsection{Benchmark algorithms, parameters setting, and datasets}
\label{sec:dataset_description}

To evaluate the predictive performance and effectiveness of images generated with LM-IGTD, we used 12 real-world and mixed-type datasets from the UCI Machine Learning Repository~\footnote{UCI Repository: http://archive.ics.uci.edu/ml/} and StatLib~\footnote{Statlib Repository: http://lib.stat.cmu.edu/datasets/}. These datasets are composed of numerical and categorical features and have between 4 and 30 total features (low-dimensional datasets). We evaluate the predictive performance of CNN-based models on images and tabular data in binary and multiclass classification scenarios. Specifically, we consider 12 low-dimensional datasets: 6 for binary classification and 6 for multiclass classification. All the datasets used are listed in Table~\ref{table:summary_datasets}, showing the number of samples, the number of categorical and numerical features and the number of classes (indicating binary and multi-class classification cases).

\begin{table}[!ht]
\centering
\caption{Datasets used in this study, indicating the total of samples, the number of features, differentiated between categorical and numerical features, and the number of classes, denoting binary and multiclass classification.}
\label{table:summary_datasets}
\scalebox{0.85}{
\begin{tabular}{|l|l|l|l|l|l|}  \hline
Datasets    & \# samples& \# features & \# categorical  & \# numerical  & \# classes  \\ 
            &           &             & features        & features      &  \\ \hline 
Crx         & 690       & 15          & 11              & 4             & 2  \\ 
Diabetes    & 520       & 16          & 15              & 1             & 2  \\ 
German      & 1000      & 20          & 17              & 3             & 2  \\ 
Hepatitis   & 155       & 19          & 18              & 1             & 2  \\ 
Ionos       & 351       & 34          & 2              & 32             & 2  \\ 
Saheart     & 462       & 9          & 1               & 8             & 2  \\  \hline
Cmc         & 1473      & 9          & 7              & 2             & 3  \\ 
Dermat      & 366       & 34          & 34              & 0             & 6  \\ 
Heart       & 303       & 13          & 8              & 5             & 5  \\ 
Annealing   & 798       & 38          & 34              & 4             & 5  \\ 
Bridges     & 106       & 10          & 9              & 1             & 3  \\             
Tae         & 151       & 5           & 2              & 3             & 3  \\ \hline
\end{tabular}
}
\end{table}

Since they are imbalanced datasets, applying standard classification ML model may not work well. To address this, a random undersampling approach was conducted, decreasing the number of images of the majority class to the number of images of the minority class. 

\subsection{Experimental setting}

Several supervised ML models were used for binary and multi-class classification scenarios for tabular data, including: the least absolute shrinkage and selection operator (LASSO), k-nearest neighbors (KNN), support vector machine (SVM) and decision tree (DT) (more details on these models are provided in~\cite{bishop2006pattern, shalev2014understanding}). To evaluate the predictive performance, data were split into a training subset (80\% samples) and a test subset (20\% samples). This was randomly performed five times by obtaining different train and test subsets, and the mean and standard deviation on these partitions are computed. To evaluate the models' performance, the mean and standard deviation of the AUCROC figure of merit were considered~\cite{shalev2014understanding}. 

To determine the values of the hyperparameters of the ML supervised models, we used $k$-fold cross-validation (CV)~\cite{bishop2006pattern} with $k=5$ and the AUCROC as the figure of merit. The following hyperparameters were explored: $C$ in the range $[1e^{-1.5}, 1e^{0.4}]$ for LASSO, $K$ between $[1, 11]$ for KNN, and $\gamma \in \{1e^{-2}, 1e^{-3}, 1e^{-4}, 1e^{-5}\}$ and $C$ in the range   $[1e^{-0.9}, 1e^{0.9}]$ for SVM. For DT, the maximum depth in the range $[2, 8]$ and the minimum samples per split were dynamically determined based on the size of the training set.

CNN-based models were implemented with the Keras framework. In pursuit of an optimal CNN architecture, we systematically explored various parameter combinations through a series of empirical experiments, leveraging the Randomized Search algorithm~\cite{bergstra2012random}. This exploration involved key architectural elements such as the number of filters, kernel size, pool size, optimizer, units in dense layer, learning rate and dropout rate. Randomized Search efficiently sampled from a wide range of parameter values, enabling a comprehensive search across the hyperparameter space~\cite{bergstra2012random}. Through these experiments, we identified a fundamental architecture that consistently exhibited superior performance across key evaluation metrics. The CNN architecture employed in our study consisted of a single convolutional layer, a max-pooling layer, a flatten layer, and a defined number of dense layers. The determination of crucial parameters, including the number of filters and the size of filters, was significantly influenced by the outcomes of the Randomized Search~\cite{bergstra2012random}. The Rectified Linear Unit (ReLu) was employed as the activation function, given its effectiveness in capturing nonlinearities. The optimization of models was conducted through the Randomized Search, selecting the most suitable optimizer from a set of options, which included Adam and RMSprop. The classification loss chosen was binary crossentropy for binary classification tasks and categorical crossentropy for multiclass classification. These loss functions are standard choices for their respective classification scenarios. The learning rate underwent dynamic updates based on the outcomes of the Randomized Search and was adjusted iteratively during training. To achieve this, a progressive reduction in the learning rate was implemented as an additional measure to enhance the model's convergence and generalization capabilities, particularly when the loss on the validation set exhibited no significant improvements. From the commencement of training, early stopping was incorporated, concluding the process if no advancement in the validation set's loss was discerned over a specified number of epochs during training.

\subsection{Evaluating noise generation for augmenting dimensionality with mixed-type features}

\begin{figure}[H]
\centering
    \begin{subfigure}{0.4\textwidth}
        \centering
        \caption*{\centering Correlation matrix with HoNG (Ionos)}
        \includegraphics[width=\linewidth]{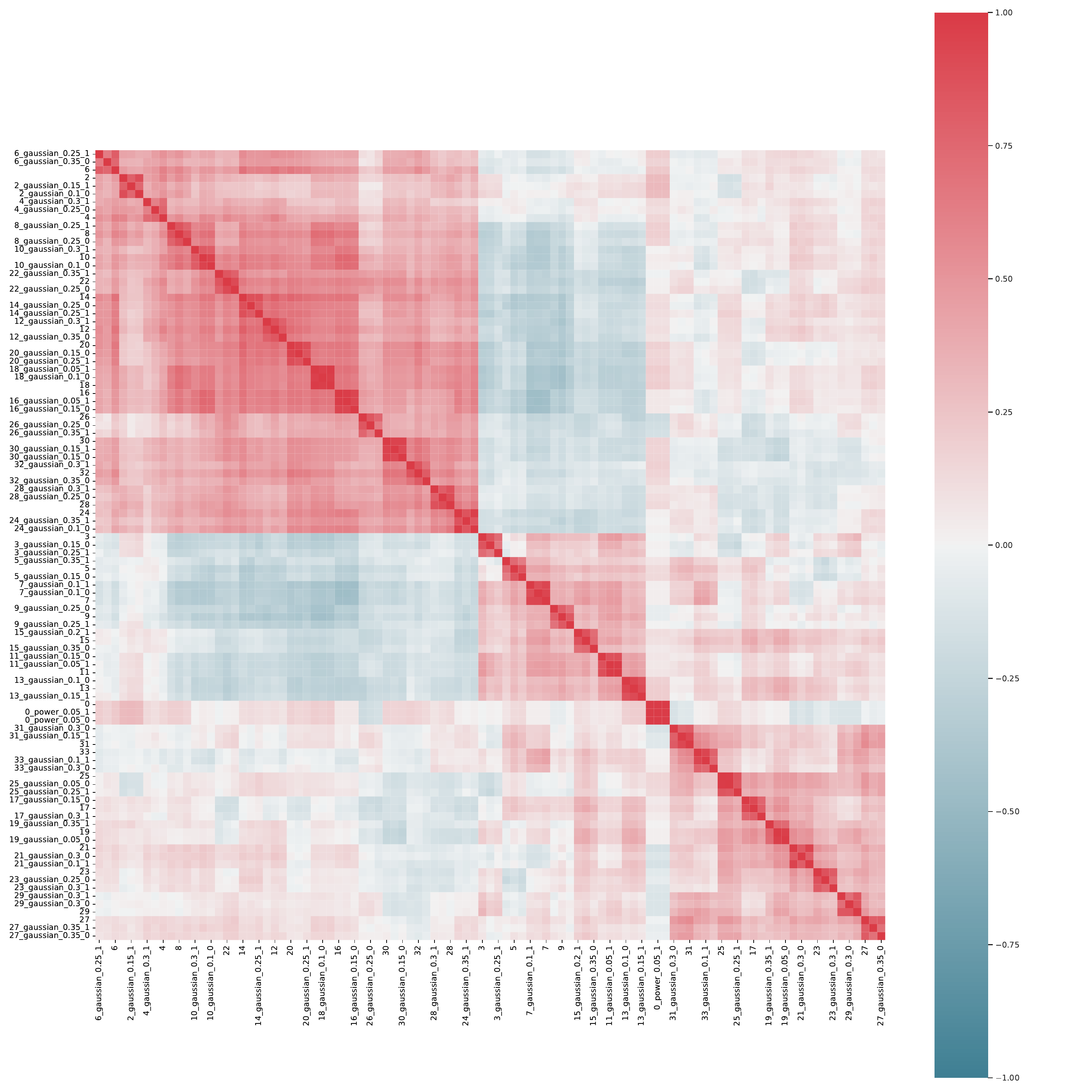}
    \end{subfigure}
    \begin{subfigure}{0.4\textwidth}
        \centering
        \caption*{Correlation matrix with HeNG (Ionos)}
        \includegraphics[width=\linewidth]{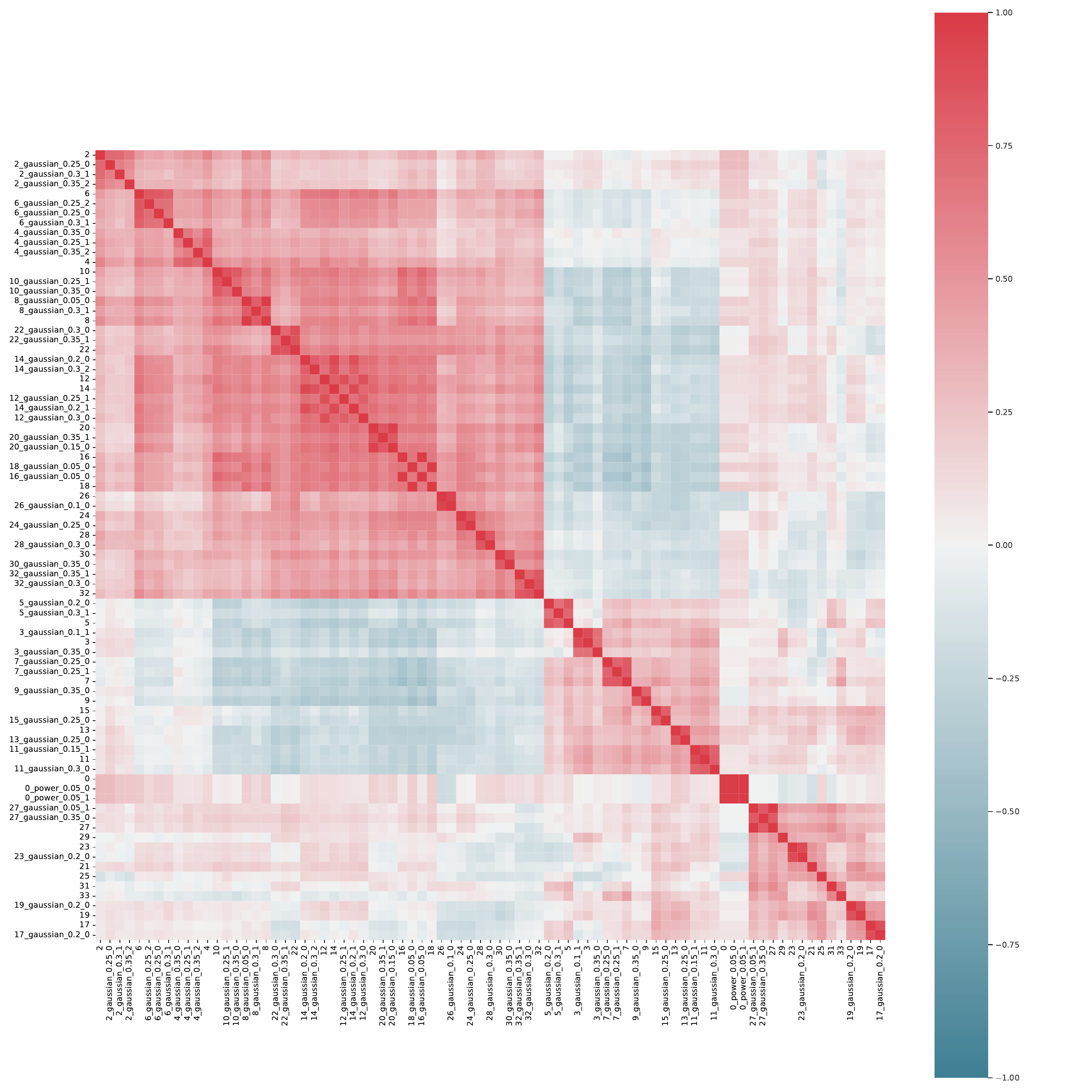}
    \end{subfigure}
    \begin{subfigure}{0.4\textwidth}
        \centering
        \caption*{Correlation matrix with HoNG (Hepatitis)}
        \includegraphics[width=\linewidth]{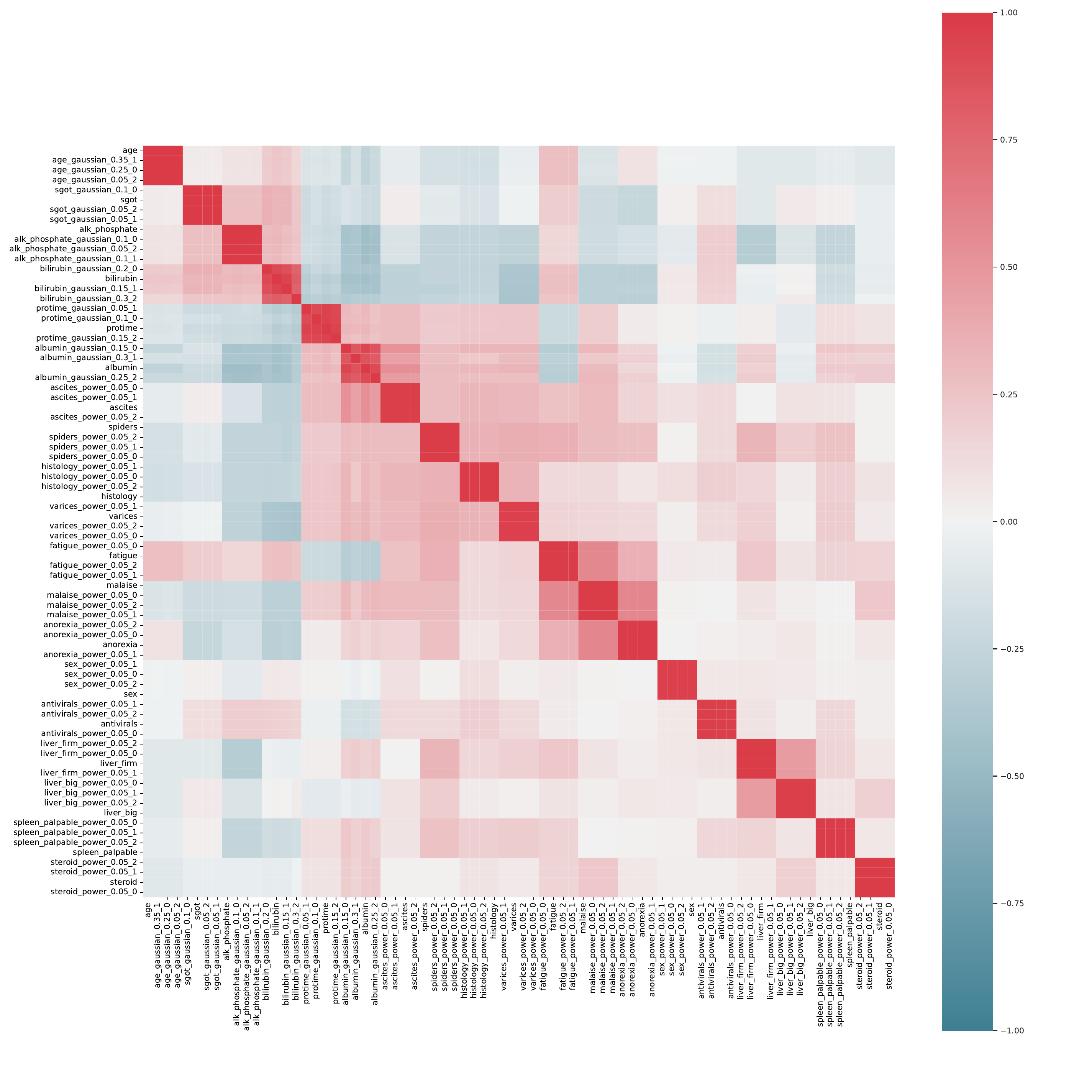}
    \end{subfigure}
    \begin{subfigure}{0.4\textwidth}
        \centering
        \caption*{Correlation matrix with HoNG (Hepatitis)}
        \includegraphics[width=\linewidth]{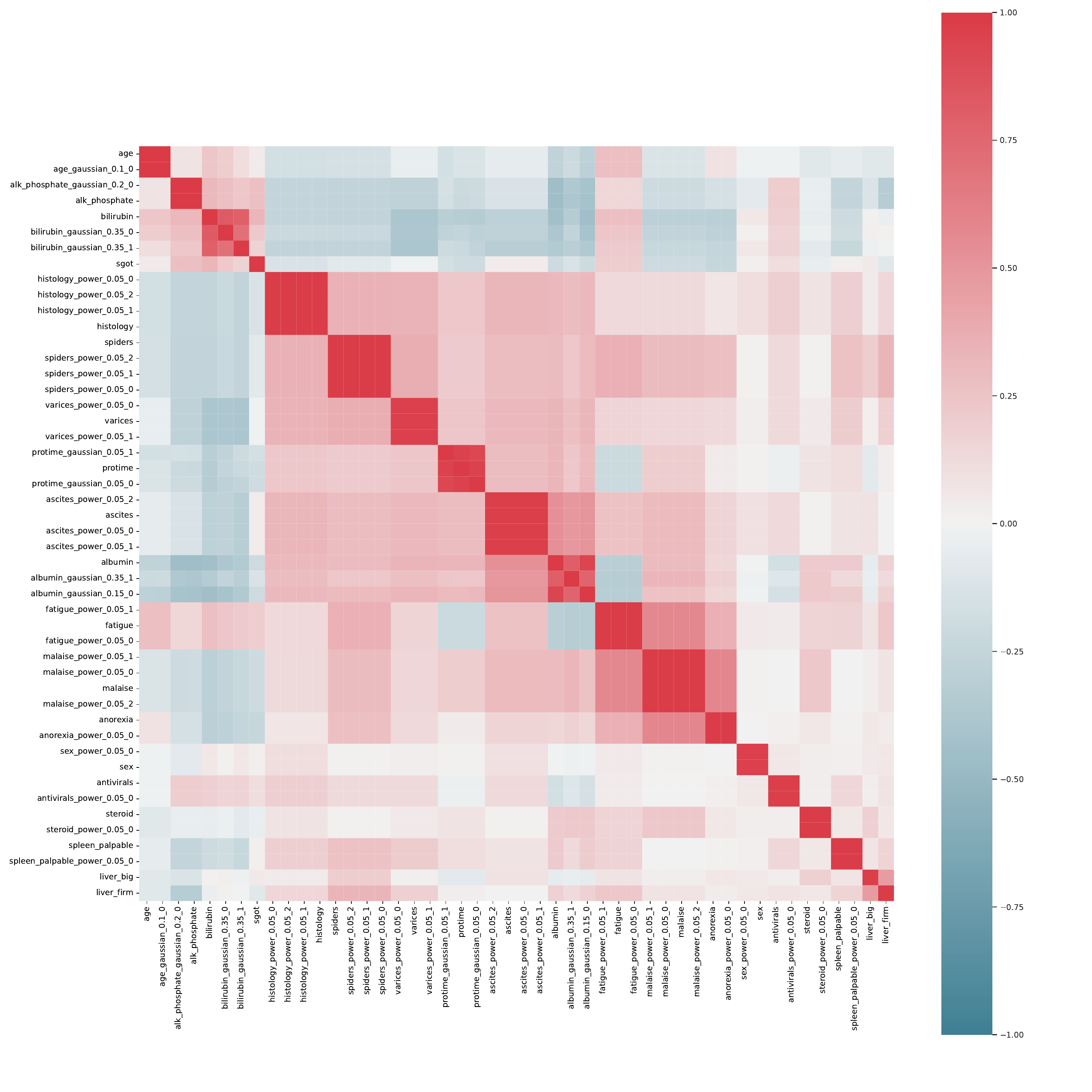}
    \end{subfigure}
    \begin{subfigure}{0.4\textwidth}
        \centering
        \caption*{Correlation matrix with HoNG (Tae)}
        \includegraphics[width=\linewidth]{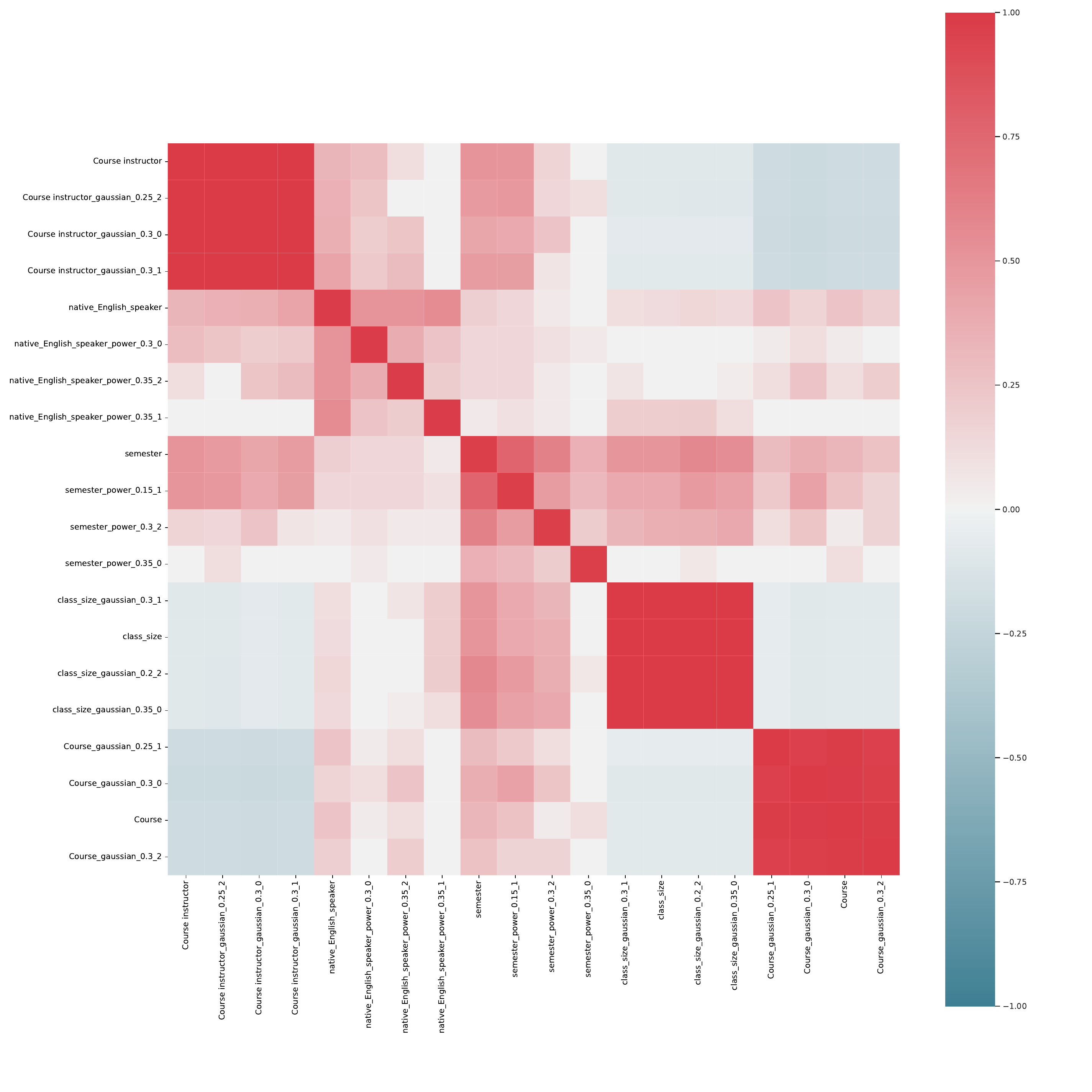}
    \end{subfigure}
        \begin{subfigure}{0.4\textwidth}
        \centering
        \caption*{Correlation matrix with HoNG (Tae)}
        \includegraphics[width=\linewidth]{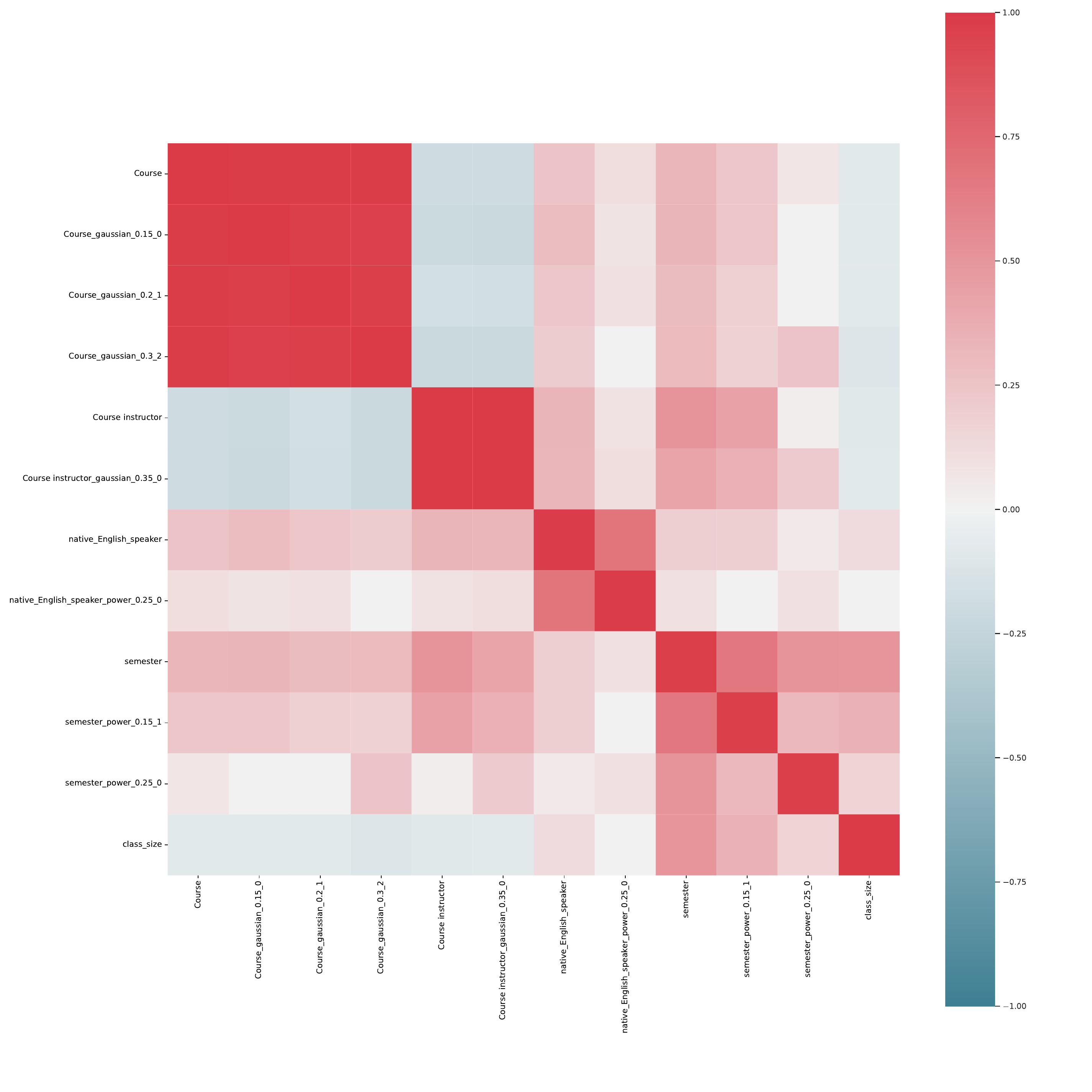}
    \end{subfigure}

    \caption{Correlation matrices between original variables and noisy variables with HoNG and HeNG in Ionos, Hepatitis and Tae datasets.}
    \label{fig:correlation_matrix}
\end{figure}

In this section, we delve into the evaluation of noise generation for augmenting dimensionality with mixed-type features. To this end, Figure~\ref{fig:correlation_matrix} shows correlation matrices of the generated noisy variables and the original variables. Specifically, correlation matrices for three selected datasets are presented. We have focused on datasets that exhibit varying characteristics in terms of the predominance of numerical or categorical features, as well as those with a balanced mix of both to illustrate different scenarios. Therefore, Figure~\ref{fig:correlation_matrix} showcases the correlation matrices for: \textit{(i)} Ionos Dataset, representing a scenario where numerical features predominate over categorical ones; \textit{(ii)} Hepatitis dataset, which represents a case where categorical features predominate over numerical ones; and \textit{(iii)} Tae dataset, representing a scenario where numerical and categorical features are balanced. This selection of datasets enabled the exploration of various scenarios, facilitating a comparison of the effects of noise generation across different dataset contexts. The generation of noisy features was carried out with the objective of enhancing the dataset's dimensionality while preserving the maximum possible correlation with the original variables. Upon examining the correlation matrices presented in Figure~\ref{fig:correlation_matrix}, it was observed that the original variables clustered closely with the corresponding noisy variables, indicating a significant relationship between them. Moreover, the correlation values between these pairs of variables were consistently high, typically ranging between 0.90 and 0.99 across the three datasets. These findings underscore the effectiveness of the noise generation process in maintaining the structural integrity of the dataset while augmenting its dimensionality.

\subsection{Classification results using tabular data versus images generated by tabular-to-image methods}

\begin{figure}[H]
\centering
\begin{subfigure}[b]{1\linewidth}
\includegraphics[width=\textwidth]{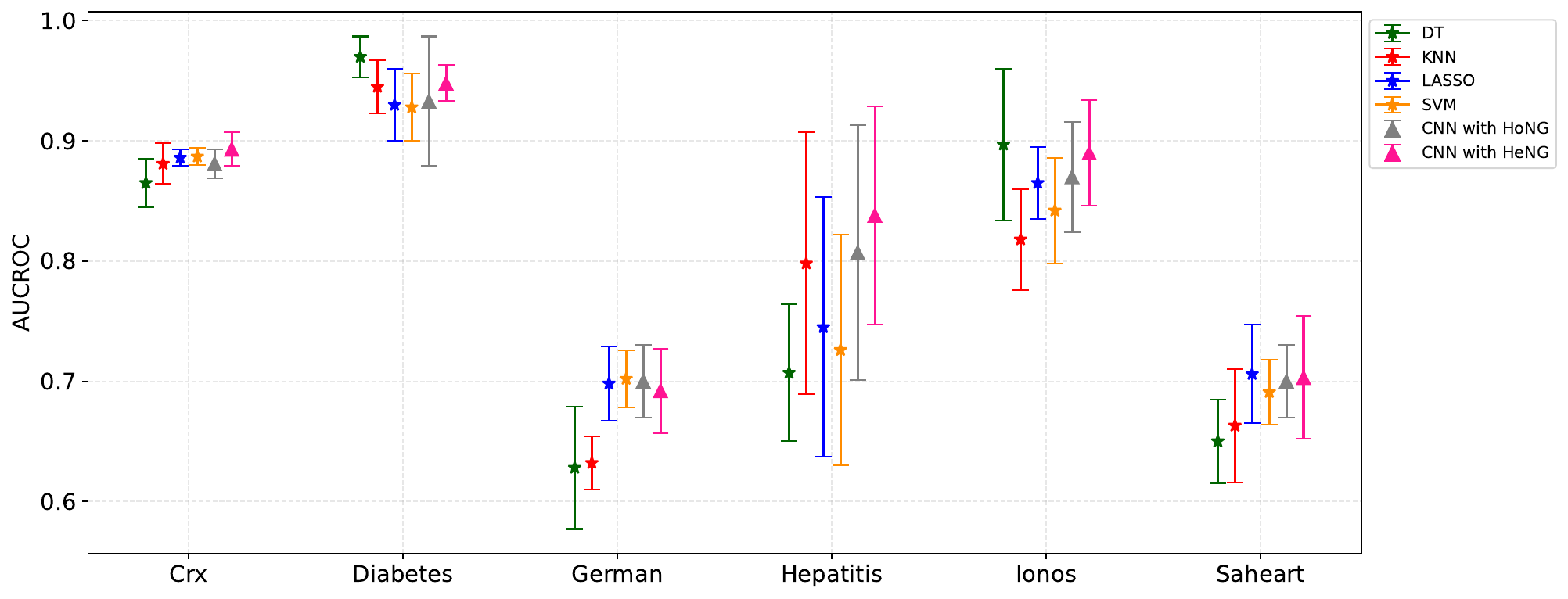}
\end{subfigure}
\caption{Mean±std of AUCROC values obtained using tabular data and LM-IGTD images and several ML and DL models for 5 test subsets of binary datasets.}
\label{fig:binary_classification_results}
\end{figure}

Figure~\ref{fig:binary_classification_results} depicts the results obtained in terms of AUCROC for six binary databases. Specifically, it showcases the outcomes of the four traditional ML models with tabular data, alongside the results of CNNs with LM-IGTD-generated images and increased dimensionality with HoNG and HeNG. This enables a direct comparison between the performance of models based on tabular data and those based on images generated by the proposed method.

In the Crx and Hepatitis datasets, LM-IGTD images and CNN models outperformed all ML models using tabular data. Similarly, across other datasets, CNN consistently outperformed at least three out of four ML models and matched the best-performing model. For instance, in the Crx dataset, the SVM model yielded the best AUCROC of 0.887 with tabular data, while CNN with HeNG for LM-IGTD images improved the performance to 0.893. Compared to tabular data, CNN showed improvements ranging from 0.6\% to 2.8\% in AUCROC for all ML models.
In the Diabetes dataset, although the DT model achieved the best AUCROC of 0.970 with tabular data, LM-IGTD images and CNN with HeNG achieved a value of 0.948, surpassing three out of four ML models. Notably, CNN showed comparable results to the DT model. LM-IGTD images with CNN consistently improved AUCROC by up to 0.3\%, 1.8\%, and 2\% in AUCROC for KNN, LASSO, and SVM models, respectively. In the German dataset, the best AUCROC of 0.702 for tabular data was achieved by the SVM model. However, CNN with HoNG for LM-IGTD images attained a slightly lower AUCROC of 0.700. Despite this marginal difference, CNN outperformed three ML models by 7.2\%, 6.8\%, and 0.2\% in AUCROC for DT, KNN, and SVM, respectively, compared to ML models for tabular data.
In the Hepatitis dataset, while KNN achieved the best AUCROC of 0.798 with tabular data, CNN using HeNG and LM-IGTD images achieved a maximum AUCROC of 0.838, surpassing all ML models. The improvement ranged from 4\% to 13.1\% in AUCROC for all ML models trained with tabular data.
Similarly, in the Ionos dataset, although DT achieved the highest AUCROC of 0.897 with tabular data, CNN surpassed three ML models with an AUCROC of 0.890. Finally, in the Saheart dataset, while LASSO achieved the best AUCROC of 0.706 with tabular data, CNN achieved comparable results of 0.703 while surpassing three ML models. Overall, LM-IGTD images trained by CNN consistently showed improvements in predictive performance compared to tabular data trained by traditional ML models.

\begin{figure}[H]
\centering
\begin{subfigure}[b]{1\linewidth}
\includegraphics[width=\textwidth]{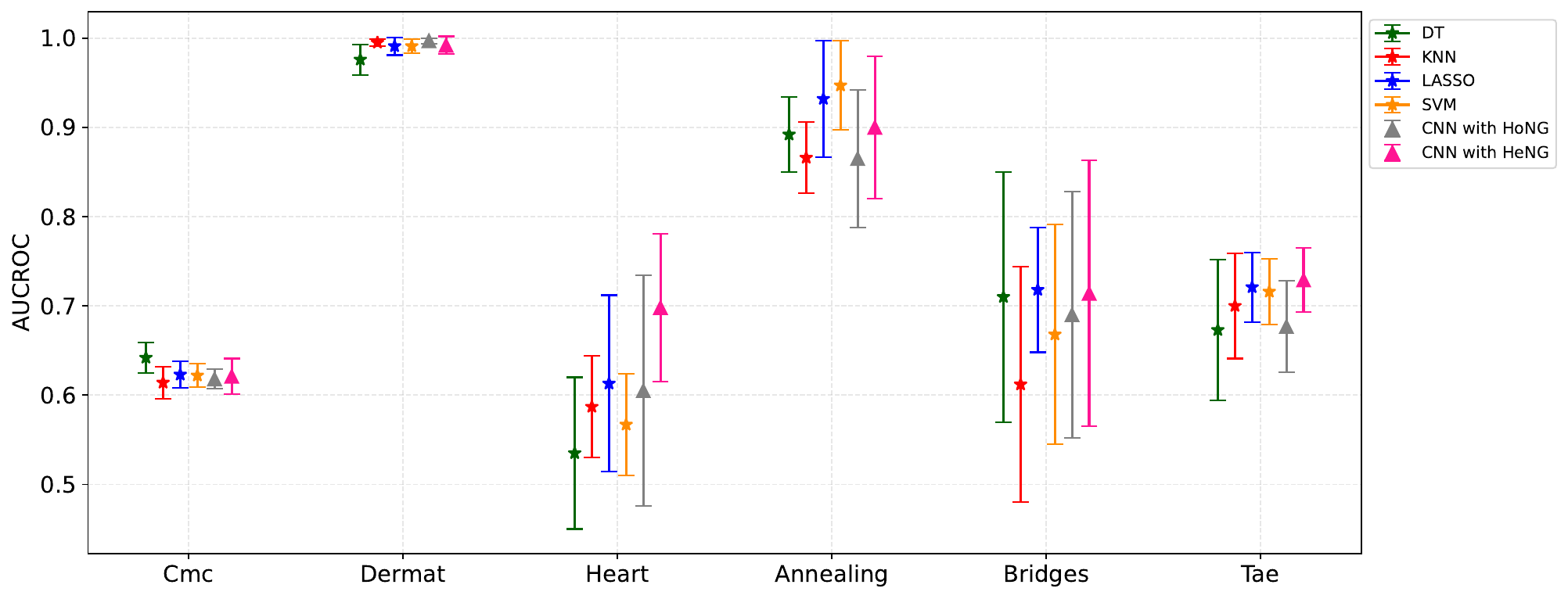}
\end{subfigure}
\caption{Mean±std of AUCROC values obtained using tabular data and LM-IGTD images and several ML and DL models for 5 test subsets of multiclass datasets.}
\label{fig:multiclass_results}
\end{figure}

Figure~\ref{fig:multiclass_results} presents the AUCROC results obtained across six multiclass databases. Specifically, in the Dermat, Heart, and Tae datasets, the CNNs models trained with LM-IGTD images exhibited superior performance compared to all ML models trained with tabular data. In contrast, in the remaining databases, the results obtained with CNNs and LM-IGTD images were quite similar to the ML models trained with tabular data.

In the Cmc dataset, the DT model achieved the best AUCROC of 0.642 with tabular data. However, utilizing LM-IGTD images with HeNG and CNN resulted in an AUCROC of 0.621, improving only one of the ML models while remaining comparable to others.
For the Dermat dataset, the KNN model achieved the highest AUCROC of 0.995 with tabular data. Nevertheless, LM-IGTD images trained by CNN surpassed all ML models, yielding an AUCROC of 0.997. Compared to tabular data, CNN showed improvements ranging from 0.2\% to 2.1\% in AUCROC for all ML models.
In the Heart dataset, the LASSO model attained the best AUCROC of 0.613 with tabular data. Conversely, using CNN with HeNG for LM-IGTD images resulted in a significantly higher AUCROC of 0.698, outperforming all ML models. The improvement achieved with CNN ranged from 8.5\% to 16.3\% in AUCROC compared to tabular data across different ML models.
In the Annealing dataset, SVM achieved the highest AUCROC of 0.947 with tabular data, while CNN with HeNG for LM-IGTD images achieved an AUCROC of 0.900. In this case, LM-IGTD images and CNN outperformed two out of four ML models, showing improvements of 0.8\% and 3.4\% in AUCROC for DT and KNN, respectively, compared to tabular data.
For the Bridges dataset, LASSO obtained the best AUCROC of 0.718 with tabular data, whereas CNN with HeNG for LM-IGTD images achieved a slightly lower AUCROC of 0.714. Nevertheless, CNN managed to surpass three out of four ML models, with improvements ranging from 0.4\% to 10.2\% in AUCROC compared to tabular data.
In the Tae dataset, LASSO achieved the highest AUCROC of 0.721 with tabular data. However, LM-IGTD images processed by CNN using HeGN achieved a superior AUCROC of 0.729, surpassing all ML models trained with tabular data. The improvement ranged from 0.8\% to 5.6\% in AUCROC for various ML models compared to tabular data.

\subsection{Interpretability analysis: feature mapping and post-hoc techniques}

\begin{figure}[H]
\centering
    \begin{subfigure}{0.57\textwidth}
        \centering
        \caption*{\centering Original image (Hepatitis)}
        \includegraphics[width=\linewidth]{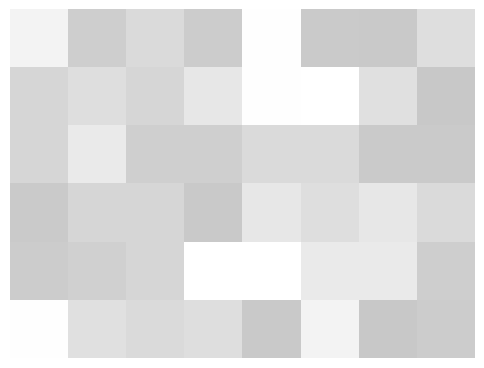}
    \end{subfigure}
    \begin{subfigure}{0.57\textwidth}
        \centering
        \caption*{Image with mapped features (Hepatitis)}
        \includegraphics[width=\linewidth]{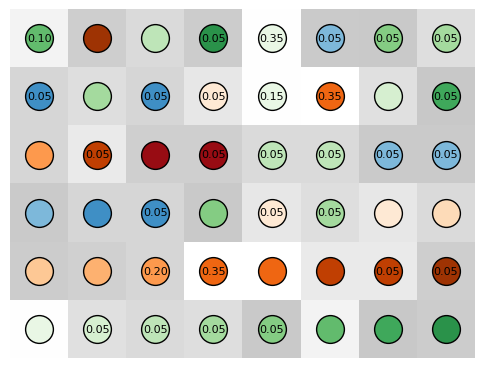}
    \end{subfigure}
           \begin{subfigure}{0.31\textwidth}
        \centering
        \caption*{Features legend (Hepatitis)}
        \includegraphics[width=\linewidth]{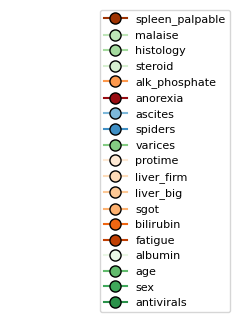}
    \end{subfigure}
    \begin{subfigure}{0.57\textwidth}
        \centering
        \caption*{Image with mapped features and Grad-CAM map (Hepatitis)}
        \includegraphics[width=\linewidth]{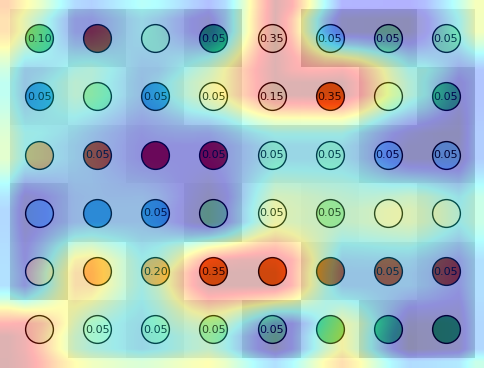}
    \end{subfigure}

    \caption{Interpretability analysis of the Hepatitis dataset: original image, feature-mapped image, and image with post-hoc Grad-CAM method}
    \label{fig:interpretability_hepatitis}
\end{figure}

\begin{figure}[H]
\centering
    \begin{subfigure}{0.4\textwidth}
        \centering
        \caption*{\centering Original image (Tae)}
        \includegraphics[width=\linewidth]{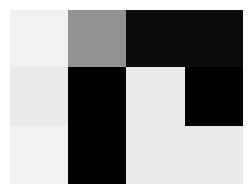}
    \end{subfigure}
    
    \begin{subfigure}{0.4\textwidth}
        \centering
        \caption*{Image with mapped features (Tae)}
        \includegraphics[width=\linewidth]{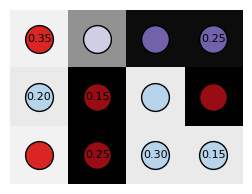}
    \end{subfigure}
           \begin{subfigure}{0.38\textwidth}
        \centering
        \caption*{Features legend (Tae)}
        \includegraphics[width=\linewidth]{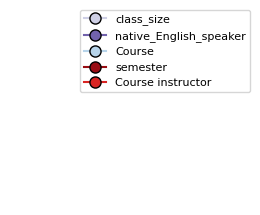}
    \end{subfigure}
    \begin{subfigure}{0.4\textwidth}
        \centering
        \caption*{Image with mapped features and Grad-CAM map (Tae)}
        \includegraphics[width=\linewidth]{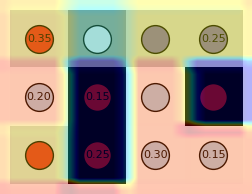}
    \end{subfigure}

    \caption{Interpretability analysis of the Tae dataset: original image, feature-mapped image, and image with post-hoc Grad-CAM method}
    \label{fig:interpretability_tae}
\end{figure}

In this section, an interpretability analysis of the images generated using the proposed LM-IGTD approach was conducted. To this end, two key techniques were explored: feature mapping and post-hoc techniques. Figure~\ref{fig:interpretability_hepatitis} and Figure~\ref{fig:interpretability_tae} illustrates this analysis using the original image, the image with mapped features, and the post-hoc Grad-CAM method, which allowed interpretability to be provided to the CNN classification process. Specifically, results are presented for two datasets that yielded the best performance: Hepatitis (binary dataset) and Tae (multiclass dataset).

Figure~\ref{fig:interpretability_hepatitis} shows the results for the Hepatitis dataset. The first subfigure showcases the image generated from the tabular data. The second subfigure illustrates the mapping of these features onto the image, with each feature color-coded and labeled alongside (where each color represents a feature). Both original and noisy features share the same color, except that the noisy features display the incorporated noise value. The third subfigure integrates the heat map generated by Grad-CAM, enhancing interpretability in the classification process. It can be observed that areas with a more orange hue correspond to those features where the network is focusing for classification, while bluer colors indicate areas less considered by the network. Specifically, for hepatitis, the network is focusing on the pixels corresponding to the features steroid, protime and bilirubin.

Figure~\ref{fig:interpretability_tae} illustrates the interpretability results for the Tae dataset. Similar to the previous dataset, the mapping and Grad-CAM techniques allowed us to identify the regions of the image where the CNN focused for classification, while also discerning which features those pixels belonged to. Specifically, for Tae, the CNN is focusing on the pixels corresponding to the feature Course.

In light of the results, the proposed method has enabled a better understanding of how the original features relate to the generated images. Furthermore, thanks to Grad-CAM, greater clarity is provided regarding the classification process carried out by the CNN.

\section{Discussion}
\label{sec:discussion}
In this study, we have addressed the challenge posed by low-dimensional and mixed-type datasets, presenting a significant advancement in this domain. While models for transforming tabular data into images have demonstrated effectiveness across various applications, their performance typically excels in high-dimensional scenarios. The number of features, directly linked to the dataset's dimensionality, critically influences the performance of most tabular-to-image methods as it dictates the resulting image size. CNN-based models necessitate images of medium to large sizes for proper training. However, achieving high dimensionality in datasets can be challenging, particularly when handling mixed-type data where the combination of numerical and categorical features introduces additional complexities.

In this context, we introduce a novel version of IGTD that integrates stochastic noise generation to address low-dimensional challenges. Moreover, our method implements specific correlations and distance measures tailored to each variable type in mixed-type datasets, enabling precise pairwise distance calculations between features. This approach not only increases the dimensionality of the dataset but also ensures the creation of representative images that improve overall performance compared to ML algorithms.

Furthermore, our variable generation method has demonstrated its capability to create variables with high levels of correlation with their original features. This underscores the effectiveness of the noise generation process in preserving the structural integrity of the dataset while augmenting its dimensionality. By maintaining strong correlations between the original features and their noisy features, our approach ensures that the generated variables capture meaningful relationships present in the data. his aspect is crucial for the creation of suitable representations in images.

Additionally, we highlight how our work addresses the inherent lack of interpretability in image generation. We have developed a mapping between the original features and the created images to identify the location of the tabular dataset features within the images. Furthermore, we have applied post hoc interpretability methods to identify relevant areas of these images and enhance interpretability in classification tasks.

\section{Conclusions}
\label{sec:conclusion}

We present a novel approach that addresses the challenges associated with low-dimensional and mixed-type datasets by introducing a new version of IGTD (LM-IGTD). LM-IGTD integrates the generation of stochastic noise and specific correlations tailored to variable types, allowing for the enhancement of dataset dimensionality while preserving the structural integrity of the data. Furthermore, in the new version of IGTD, we have implemented the mapping of original features to images to maintain interpretability and visualize the spatial distribution of features within the images. Additionally, through methods such as Grad-CAM, we have provided interpretability to the classification task performed by the CNN.

Evaluation across multiple datasets has shown the robustness and generalization capability of this approach. In particular, our method outperformed traditional ML models trained on tabular data in five out of twelve datasets when using images generated with LM-IGTD and CNN. In the remaining datasets, we consistently surpassed three out of four traditional ML models, achieving similar results to the fourth model. These findings underscore the effectiveness of our approach in generating meaningful representations from tabular data, surpassing or matching the performance of traditional ML models in various contexts. Our method not only addresses the challenges posed by low-dimensional and mixed-type datasets but also provides new opportunities for the development of methodologies capable of handling various types of data, facilitating the transformation of tabular data into images.

\bibliographystyle{vancouver}
\bibliography{literature}

\begin{thebibliography}{10}

\bibitem{borisov2022deep}
Borisov V, Leemann T, Se{\ss}ler K, Haug J, Pawelczyk M, Kasneci G.
\newblock Deep neural networks and tabular data: A survey.
\newblock IEEE Transactions on Neural Networks and Learning Systems. 2022.

\bibitem{oord2016wavenet}
Oord Avd, Dieleman S, Zen H, Simonyan K, Vinyals O, Graves A, et~al.
\newblock Wavenet: A generative model for raw audio.
\newblock arXiv preprint arXiv:160903499. 2016.

\bibitem{devlin2018bert}
Devlin J, Chang MW, Lee K, Toutanova K.
\newblock Bert: Pre-training of deep bidirectional transformers for language understanding.
\newblock arXiv preprint arXiv:181004805. 2018.

\bibitem{he2016deep}
He K, Zhang X, Ren S, Sun J.
\newblock Deep residual learning for image recognition.
\newblock In: Proceedings of the IEEE conference on computer vision and pattern recognition; 2016. p. 770-8.

\bibitem{lauriola2022introduction}
Lauriola I, Lavelli A, Aiolli F.
\newblock An introduction to deep learning in natural language processing: Models, techniques, and tools.
\newblock Neurocomputing. 2022;470:443-56.

\bibitem{zhang2015deep}
Zhang W, Li R, Deng H, Wang L, Lin W, Ji S, et~al.
\newblock Deep convolutional neural networks for multi-modality isointense infant brain image segmentation.
\newblock NeuroImage. 2015;108:214-24.

\bibitem{wu2015scalable}
Wu G, Kim M, Wang Q, Munsell BC, Shen D.
\newblock Scalable high-performance image registration framework by unsupervised deep feature representations learning.
\newblock IEEE transactions on biomedical engineering. 2015;63(7):1505-16.

\bibitem{suk2014hierarchical}
Suk HI, Lee SW, Shen D, Initiative ADN, et~al.
\newblock Hierarchical feature representation and multimodal fusion with deep learning for AD/MCI diagnosis.
\newblock NeuroImage. 2014;101:569-82.

\bibitem{krizhevsky2017imagenet}
Krizhevsky A, Sutskever I, Hinton GE.
\newblock ImageNet classification with deep convolutional neural networks.
\newblock Communications of the ACM. 2017;60(6):84-90.

\bibitem{anwar2018medical}
Anwar SM, Majid M, Qayyum A, Awais M, Alnowami M, Khan MK.
\newblock Medical image analysis using convolutional neural networks: a review.
\newblock Journal of medical systems. 2018;42:1-13.

\bibitem{arik2021tabnet}
Arik S{\"O}, Pfister T.
\newblock Tabnet: Attentive interpretable tabular learning.
\newblock In: Proceedings of the AAAI conference on artificial intelligence. vol.~35; 2021. p. 6679-87.

\bibitem{medeiros2023comparative}
Medeiros~Neto L, Rogerio~da Silva~Neto S, Endo PT.
\newblock A comparative analysis of converters of tabular data into image for the classification of Arboviruses using Convolutional Neural Networks.
\newblock Plos one. 2023;18(12):e0295598.

\bibitem{shwartz2022tabular}
Shwartz-Ziv R, Armon A.
\newblock Tabular data: Deep learning is not all you need.
\newblock Information Fusion. 2022;81:84-90.

\bibitem{grinsztajn2022tree}
Grinsztajn L, Oyallon E, Varoquaux G.
\newblock Why do tree-based models still outperform deep learning on typical tabular data?
\newblock Advances in Neural Information Processing Systems. 2022;35:507-20.

\bibitem{bragilovski2023tltd}
Bragilovski M, Kapri Z, Rokach L, Levy-Tzedek S.
\newblock TLTD: Transfer Learning for Tabular Data.
\newblock Applied Soft Computing. 2023;147:110748.

\bibitem{wolf2022daft}
Wolf TN, P{\"o}lsterl S, Wachinger C, Initiative ADN, et~al.
\newblock DAFT: a universal module to interweave tabular data and 3D images in CNNs.
\newblock NeuroImage. 2022;260:119505.

\bibitem{lee2022tab2vox}
Lee E, Nam M, Lee H.
\newblock Tab2vox: CNN-based multivariate multilevel demand forecasting framework by tabular-to-voxel image conversion.
\newblock Sustainability. 2022;14(18):11745.

\bibitem{sharma2019deepinsight}
Sharma A, Vans E, Shigemizu D, Boroevich KA, Tsunoda T.
\newblock DeepInsight: A methodology to transform a non-image data to an image for convolution neural network architecture.
\newblock Scientific reports. 2019;9(1):11399.

\bibitem{bazgir2020representation}
Bazgir O, Zhang R, Dhruba SR, Rahman R, Ghosh S, Pal R.
\newblock Representation of features as images with neighborhood dependencies for compatibility with convolutional neural networks.
\newblock Nature communications. 2020;11(1):4391.

\bibitem{zhu2021converting}
Zhu Y, Brettin T, Xia F, Partin A, Shukla M, Yoo H, et~al.
\newblock Converting tabular data into images for deep learning with convolutional neural networks.
\newblock Scientific reports. 2021;11(1):11325.

\bibitem{zhou2016learning}
Zhou B, Khosla A, Lapedriza A, Oliva A, Torralba A.
\newblock Learning deep features for discriminative localization.
\newblock In: Proceedings of the IEEE conference on computer vision and pattern recognition; 2016. p. 2921-9.

\bibitem{damri2023towards}
Damri A, Last M, Cohen N.
\newblock Towards efficient image-based representation of tabular data.
\newblock Neural Computing and Applications. 2023:1-21.

\bibitem{hashemi2019enlarging}
Hashemi M.
\newblock Enlarging smaller images before inputting into convolutional neural network: zero-padding vs. interpolation.
\newblock Journal of Big Data. 2019;6(1):1-13.

\bibitem{jia2023scdeepinsight}
Jia S, Lysenko A, Boroevich KA, Sharma A, Tsunoda T.
\newblock scDeepInsight: a supervised cell-type identification method for scRNA-seq data with deep learning.
\newblock Briefings in Bioinformatics. 2023;24(5):bbad266.

\bibitem{han2019convolutional}
Han H, Li Y, Zhu X.
\newblock Convolutional neural network learning for generic data classification.
\newblock Information Sciences. 2019;477:448-65.

\bibitem{sharma2021deepfeature}
Sharma A, Lysenko A, Boroevich KA, Vans E, Tsunoda T.
\newblock DeepFeature: feature selection in nonimage data using convolutional neural network.
\newblock Briefings in Bioinformatics. 2021;22(6):bbab297.

\bibitem{sharma2022classification}
Sharma A, Kumar D.
\newblock Classification with 2-D Convolutional Neural Networks for breast cancer diagnosis.
\newblock Scientific Reports. 2022;12(1):21857.

\bibitem{sisodia2022feature}
Sisodia D, Sisodia DS.
\newblock Feature space transformation of user-clicks and deep transfer learning framework for fraudulent publisher detection in online advertising.
\newblock Applied Soft Computing. 2022;125:109142.

\bibitem{iqbal2022dynamic}
Iqbal MI, Mukta MSH, Hasan AR, Islam S.
\newblock A dynamic weighted tabular method for convolutional neural networks.
\newblock IEEE Access. 2022;10:134183-98.

\bibitem{matsuda2023hacnet}
Matsuda T, Uchida K, Saito S, Shirakawa S.
\newblock HACNet: End-to-end learning of interpretable table-to-image converter and convolutional neural network.
\newblock Knowledge-Based Systems. 2023:111293.

\bibitem{briner2023tabular}
Briner N, Cullen D, Halladay J, Miller D, Primeau R, Avila A, et~al.
\newblock Tabular-to-Image Transformations for the Classification of Anonymous Network Traffic using Deep Residual Networks.
\newblock IEEE Access. 2023.

\bibitem{tang2022vec2image}
Tang H, Yu X, Liu R, Zeng T.
\newblock Vec2image: an explainable artificial intelligence model for the feature representation and classification of high-dimensional biological data by vector-to-image conversion.
\newblock Briefings in Bioinformatics. 2022;23(2):bbab584.

\bibitem{castillo2023tinto}
Castillo-Cara M, Talla-Chumpitaz R, Garc{\'\i}a-Castro R, Orozco-Barbosa L.
\newblock TINTO: Converting Tidy Data into image for classification with 2-Dimensional Convolutional Neural Networks.
\newblock SoftwareX. 2023;22:101391.

\bibitem{zandavi2023fotomics}
Zandavi SM, Liu D, Chung V, Anaissi A, Vafaee F.
\newblock Fotomics: fourier transform-based omics imagification for deep learning-based cell-identity mapping using single-cell omics profiles.
\newblock Artificial Intelligence Review. 2023;56(7):7263-78.

\bibitem{li2021survey}
Li Z, Liu F, Yang W, Peng S, Zhou J.
\newblock A survey of convolutional neural networks: analysis, applications, and prospects.
\newblock IEEE transactions on neural networks and learning systems. 2021.

\bibitem{yang2022unbox}
Yang G, Ye Q, Xia J.
\newblock Unbox the black-box for the medical explainable AI via multi-modal and multi-centre data fusion: A mini-review, two showcases and beyond.
\newblock Information Fusion. 2022;77:29-52.

\bibitem{vilone2021notions}
Vilone G, Longo L.
\newblock Notions of explainability and evaluation approaches for explainable artificial intelligence.
\newblock Information Fusion. 2021;76:89-106.

\bibitem{selvaraju2017grad}
Selvaraju RR, Cogswell M, Das A, Vedantam R, Parikh D, Batra D.
\newblock Grad-cam: Visual explanations from deep networks via gradient-based localization.
\newblock In: Proceedings of the IEEE international conference on computer vision; 2017. p. 618-26.

\bibitem{ali2023explainable}
Ali S, Abuhmed T, El-Sappagh S, Muhammad K, Alonso-Moral JM, Confalonieri R, et~al.
\newblock Explainable Artificial Intelligence (XAI): What we know and what is left to attain Trustworthy Artificial Intelligence.
\newblock Information Fusion. 2023;99:101805.

\bibitem{chen2022noise}
Chen H, Birkelund Y, Batalden BM, Barabadi A.
\newblock Noise-intensification data augmented machine learning for day-ahead wind power forecast.
\newblock Energy Reports. 2022;8:916-22.

\bibitem{momeny2021learning}
Momeny M, Neshat AA, Hussain MA, Kia S, Marhamati M, Jahanbakhshi A, et~al.
\newblock Learning-to-augment strategy using noisy and denoised data: Improving generalizability of deep CNN for the detection of COVID-19 in X-ray images.
\newblock Computers in Biology and Medicine. 2021;136:104704.

\bibitem{vincent2010stacked}
Vincent P, Larochelle H, Lajoie I, Bengio Y, Manzagol PA, Bottou L.
\newblock Stacked denoising autoencoders: Learning useful representations in a deep network with a local denoising criterion.
\newblock Journal of Machine Learning Research. 2010;11(12).

\bibitem{chushig2022learning}
Chushig-Muzo D, Soguero-Ruiz C, Miguel~Bohoyo Pd, Mora-Jim{\'e}nez I, et~al.
\newblock Learning and visualizing chronic latent representations using electronic health records.
\newblock BioData Mining. 2022;15(1):1-27.

\bibitem{remeseiro2019review}
Remeseiro B, Bolon-Canedo V.
\newblock A review of feature selection methods in medical applications.
\newblock Computers in biology and medicine. 2019;112:103375.

\bibitem{pes2020ensemble}
Pes B.
\newblock Ensemble feature selection for high-dimensional data: a stability analysis across multiple domains.
\newblock Neural Computing and Applications. 2020;32(10):5951-73.

\bibitem{seijo2017ensemble}
Seijo-Pardo B, Porto-D{\'\i}az I, Bol{\'o}n-Canedo V, Alonso-Betanzos A.
\newblock Ensemble feature selection: homogeneous and heterogeneous approaches.
\newblock Knowledge-Based Systems. 2017;118:124-39.

\bibitem{yu2022robust}
Yu H, Hutson AD.
\newblock A robust Spearman correlation coefficient permutation test.
\newblock Communications in Statistics-Theory and Methods. 2022:1-13.

\bibitem{kornbrot2014point}
Kornbrot D.
\newblock Point biserial correlation.
\newblock Wiley StatsRef: Statistics Reference Online. 2014.

\bibitem{baak2020new}
Baak M, Koopman R, Snoek H, Klous S.
\newblock A new correlation coefficient between categorical, ordinal and interval variables with Pearson characteristics.
\newblock Computational Statistics \& Data Analysis. 2020;152:107043.

\bibitem{bishop2006pattern}
Bishop CM.
\newblock Pattern Recognition and Machine Learning (Information Science and Statistics).
\newblock Berlin, Heidelberg: Springer-Verlag; 2006.

\bibitem{shalev2014understanding}
Shalev-Shwartz S, Ben-David S.
\newblock Understanding machine learning: From theory to algorithms.
\newblock Cambridge university press; 2014.

\bibitem{bergstra2012random}
Bergstra J, Bengio Y.
\newblock Random search for hyper-parameter optimization.
\newblock Journal of machine learning research. 2012;13(2).

\end{thebibliography}

\end{document}